\documentclass[letterpaper, 10 pt, conference]{ieeeconf}  

\IEEEoverridecommandlockouts                              

\usepackage{amsmath,amssymb,amsfonts}
\usepackage{algorithmic}
\usepackage{multirow}
\usepackage{mathtools}
\usepackage{graphicx}
\usepackage{textcomp}
\usepackage{hyperref}
\usepackage{xcolor}
\usepackage{booktabs}
\usepackage{ dsfont }
\usepackage{caption}
\usepackage{subcaption}
\usepackage{xcolor} 
\usepackage{pdfpages}
\usepackage{graphicx}
\usepackage{pgf}
\usepackage{import}   

\def\BibTeX{{\rm B\kern-.05em{\sc i\kern-.025em b}\kern-.08em
    T\kern-.1667em\lower.7ex\hbox{E}\kern-.125emX}}

\title{\LARGE \bf
Robust RL Control for Bipedal Locomotion with Closed Kinematic Chains
}
\author{
Egor Maslennikov$^{1,2}$, 
Eduard Zaliaev$^{1,3}$, 
Nikita Dudorov$^{1}$, 
Oleg Shamanin$^{1}$, 
Karanov Dmitry$^{1}$, \\
Gleb Afanasev$^{1}$, 
Alexey Burkov$^{1}$, 
Egor Lygin$^{1,4}$, 
Simeon Nedelchev$^{1,3}$, 
Evgeny Ponomarev$^{1}$
\thanks{$^{1}$ Sber Robotics Center, Moscow, Russia, {\tt\footnotesize  \{nadudorov, omakshamanin,dpkaranov,giafanasyev, burkov.a.m,eserponomarev\}@sberbank.ru}}
\thanks{$^{2}$ Moscow Institute of Physics and Technology (MIPT), Dolgoprudny,  Russia, {\tt\footnotesize  maslennikov.em@phystech.edu}}%
\thanks{$^{3}$ Innopolis University, Innopolis, Russia. {\tt\footnotesize s.nedelchev@innopolis.university, e.zalyaev@innopolis.ru}}%
\thanks{$^{4}$ ITMO University, Saint Petersburg, Russia, {\tt\footnotesize edlygin@itmo.ru}}%
}

\begin{document}

\maketitle
\thispagestyle{empty}
\pagestyle{empty}


\begin{abstract}
Developing robust locomotion controllers for bipedal robots with closed kinematic chains presents unique challenges, particularly since most reinforcement learning (RL) approaches simplify these parallel mechanisms into serial models during training. We demonstrate that this simplification significantly impairs sim-to-real transfer by failing to capture essential aspects such as joint coupling, friction dynamics, and motor-space control characteristics. In this work, we present an RL framework that explicitly incorporates closed-chain dynamics and validate it on our custom-built robot TopA. Our approach enhances policy robustness through symmetry-aware loss functions, adversarial training, and targeted network regularization. Experimental results demonstrate that our integrated approach achieves stable locomotion across diverse terrains, significantly outperforming methods based on simplified kinematic models.
\end{abstract}


\section{Introduction}

Reinforcement learning (RL) has fundamentally transformed robotic locomotion control, demonstrating unprecedented success in handling complex, high-dimensional control tasks \cite{disney_robot, wtw, revisit_rwd_design_cassie}. The key advantage of RL lies in its ability to learn robust policies through domain randomization, enabling adaptation to real-world uncertainties without requiring explicit modeling of all environmental factors. This data-driven approach is particularly valuable for bipedal locomotion, where traditional model-based methods often struggle with the complexity of dynamic interactions and environmental variability.

However, the application of RL to physical bipedal robots reveals a critical gap between simulation and reality, particularly for systems with closed kinematic chains. These parallel mechanisms, common in practical bipedal designs, serve essential functions: they provide mechanical advantage, reduce actuator requirements, and enhance structural stability. Despite their prevalence, most RL frameworks default to serial robot models, sacrificing physical accuracy for computational efficiency. This simplification is evident in work with several prominent platforms, including Berkeley Humanoid \cite{berkeley_humanoid}, GR-1 \cite{gr1_rl}, and G1 \cite{exbody2}. While researchers have proposed various approaches, such as virtual approximations \cite{digit_rl} or specialized but computationally intensive simulators \cite{cassie_rl}, these methods address symptoms rather than resolving the fundamental challenge of accurately modeling closed kinematic chains.

In parallel with these modeling challenges, the RL community has made significant advances in policy learning techniques with direct applications to bipedal locomotion. Notable developments include leveraging morphological symmetry for more natural gaits \cite{ordonez2023discrete}, adversarial training for disturbance rejection \cite{adversarial}, and policy network constraints for enhanced stability \cite{radosavovic2024learning}. Yet these promising techniques have primarily been developed and validated either in isolation or within simplified serial model frameworks, leaving a significant gap in understanding their effectiveness when applied to parallel mechanisms.

The intersection of these two research directions—accurate modeling of closed kinematic chains and advanced policy learning techniques—presents a compelling frontier for advancing bipedal locomotion control. Addressing this intersection is particularly critical for effective sim-to-real transfer, where the combination of physically accurate models with robust policy learning methods could dramatically improve real-world performance.

\begin{figure}
    \centerline{\includegraphics[scale=0.165]{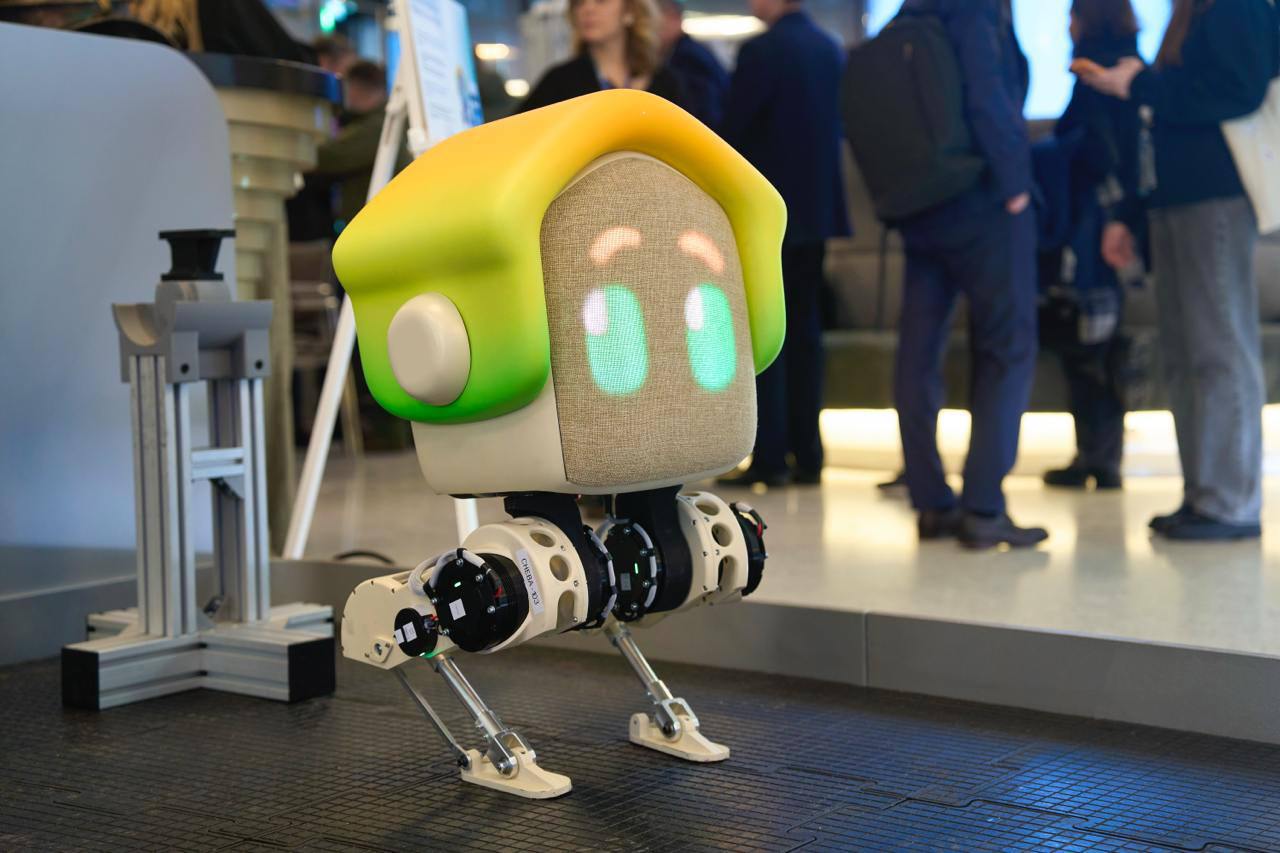}}
    \caption{The TopA bipedal robot.}
    \label{fig:cheba_depiction_live}
    \vspace*{-5mm} 
\end{figure}

To address these challenges, we present a comprehensive approach that integrates accurate modeling of closed kinematic chains with state-of-the-art policy learning techniques. Our contributions to the field of RL-based bipedal locomotion control include:

\begin{itemize}
    \item A RL training pipeline that explicitly incorporates closed kinematic chains, achieving superior sim-to-real transfer through accurate modeling of joint coupling, friction dynamics, and motor-space control characteristics.
    \item An integrated robustness enhancement approach combining symmetry-aware learning, adversarial training, and structured policy regularization—demonstrating that these techniques remain effective and complementary when applied to parallel mechanisms.
    \item Validation on TopA (Fig. \ref{fig:cheba_depiction_live}), a custom bipedal robot designed specifically to study parallel mechanism control, achieving reliable locomotion across varied terrains and extended operation in human environments.
    \item Comprehensive experimental analysis through systematic ablation studies and robustness evaluations, quantifying the benefits of the integrated approach compared to conventional methods.
\end{itemize}

The remainder of this paper is organized as follows: In Section II, we detail our learning framework, beginning with the formal problem formulation and proceeding to our specific implementation choices for training methodology and robustness enhancement techniques. Section III presents the TopA robot platform, describing its hardware design and the specific challenges posed by its closed kinematic chains and provides experimental validation through ablation studies and robustness evaluations, demonstrating the effectiveness of our approach in both simulated and real-world environments. Finally, Section IV concludes with a discussion of our results and their implications for future research in robust bipedal locomotion.

\section{Learning Pipeline}

Our learning pipeline addresses the fundamental challenge of accurately modeling and controlling bipedal robots with closed kinematic chains. Through the integration of precise physical modeling with advanced policy learning techniques, we create a framework that significantly narrows the sim-to-real gap for these complex systems.

\subsection{Problem formulation}
The problem we solve is a decision process on $\mathcal{M} = \langle \mathcal{S}, \mathcal{A}, T, R, \gamma \rangle$ with discounted reward maximization objective. $\mathcal{S}$ is the state space, $\mathcal{A}$ is the action space, $T: \mathcal{S} \times \mathcal{A} \to \mathcal{S}$ is the transition function, $R: \mathcal{S} \times \mathcal{A} \to \mathbb{R}$ is the reward function, $\gamma \in (0, 1]$ is the discount factor. In our case $T$ is a deterministic function: $s_{t+1} = T(s_t, a_t)$. Rewards are received according to selected actions $a_t$: $r_t = R(s_t, a_t)$. The actions are chosen according to policy $\pi$ which is a conditional distribution of action given observed state $o_t = f(s_t)$: $\pi(a_t | o_t)$. The policy is parameterized by a neural network with parameters $\theta: \pi = \pi_{\theta}$, which are optimized with the objective: 
\begin{equation}
\theta^* = \arg\max_{\theta} \mathbb{E}_{\tau \sim \pi_{\theta}} \left[ \sum_{t=0}^{\infty} \gamma^t r_t \right]
\end{equation}

\subsection{Training methodology}
With this formal framework established, we now detail our implementation choices for training an effective locomotion policy.

\begin{figure*}[t]
   \centering
   \includegraphics[width=145mm]{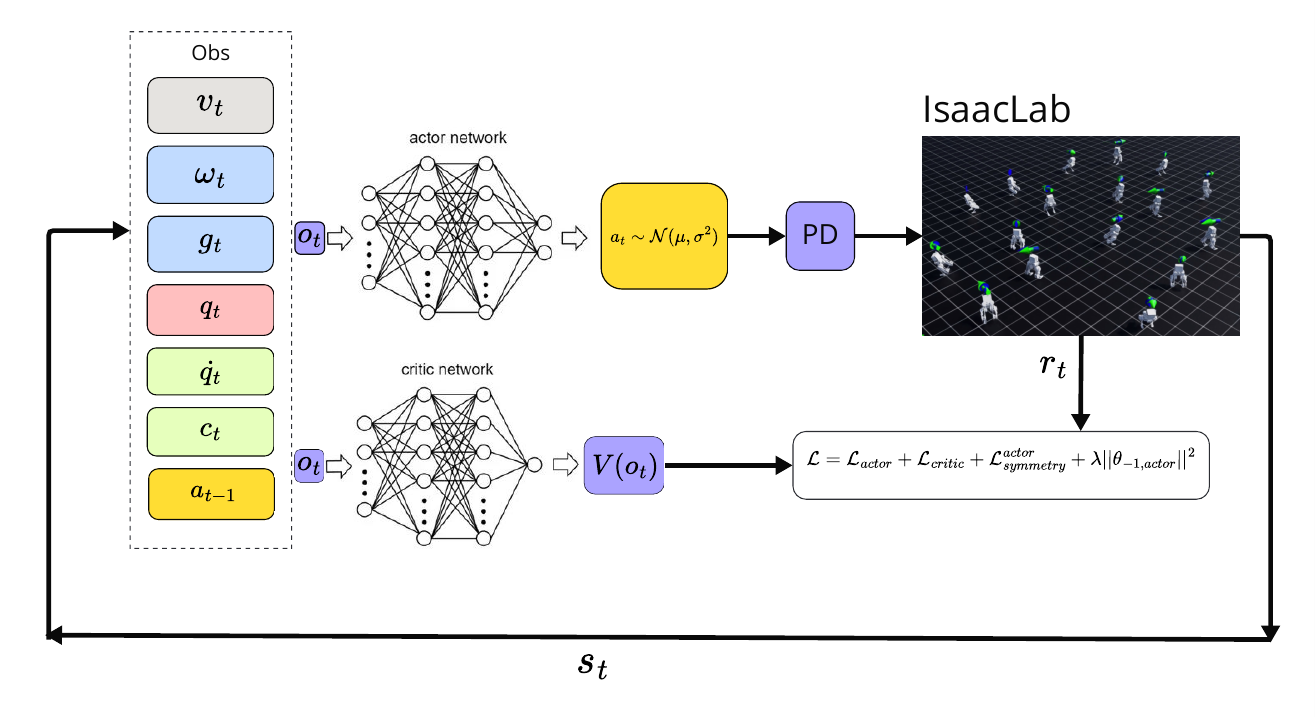}
   \caption{Policy training scheme.}
   \label{fig:train_scheme}
   \vspace{-3mm}
\end{figure*}

We implement the policy $\pi$ using PPO in an Actor-Critic architecture \cite{ppo}. Both actor and critic networks utilize Multi-layer perceptron (MLP) architectures, with the actor serving as the primary locomotion control policy $\pi$. Both networks share identical input consisting of robot proprioception:
$$\mathbf{o_t} = \left[ \mathbf{v_t}, \boldsymbol{\omega_t}, \mathbf{g_t}, \mathbf{q_t}, \mathbf{\dot{q}_t},  \mathbf{a_{t - 1}}, \mathbf{c_t}\right], $$ 
where $\mathbf{v_t}$ and $\boldsymbol{\omega_t}$ represent linear and angular velocities in the robot frame, $\mathbf{g_t}$ denotes the gravity vector in the robot frame, $\mathbf{q_t}$ and $\mathbf{\dot{q}_t}$ are motor positions and velocities, $\mathbf{a_{t - 1}}$ represents the previous action, and $\mathbf{c_t}$ contains the command vector. The command vector $\mathbf{c_t}$ comprises $v_x$, $v_y$, and $w_z$, which specify the desired linear velocities in $X$ and $Y$ directions and angular velocity around the $Z$ axis, respectively, all expressed in the local frame.

\begin{table}[h]
    \centering
    \caption{PPO Parameters}
    \begin{tabular}{l|l}
        \toprule
         PPO clip range & 0.2 \\
         GAE $\lambda$ & 0.95 \\
         Learning rate & 3e-5 \\
         Reward discount factor $\gamma$ & 0.99 \\
         Max gradient norm & 0.5 \\
         Desired KL-divergence & 0.02 \\
         Number of environments & 4096 \\
         Number of environment steps per training batch & 24 \\
         Learning epochs per training batch & 5 \\
         Number of mini-batches per training batch & 4 \\
         Actor layer dimensions & [512, 256, 128] \\
         Critic layer dimensions & [512, 256, 128] \\
         Adversary layer dimensions & [512, 256, 128] \\
         Activation function & ELU \\
         \bottomrule
    \end{tabular}
    \label{tab:ppo}
\end{table}

\begin{table}[h]
    \centering
    \caption{Rewards}
    \begin{tabular}{llr}
    \toprule Reward term & Equation & Weight \\
    \midrule Lin. velocity tracking & $\exp \left\{- \frac{\|\mathbf{v}_{x y}^{\text {cmd }}-\mathbf{v}_{x y}\|_2^2}{\sigma_v}  \right\}$ & 3.0 \\
    Ang. velocity tracking & $\exp \left\{- \frac{\left(\omega_z^{\text {cmd }}-\omega_z \right)^2}{\sigma_\omega} \right\}$ & 1.5 \\
    Linear velocity $(z)$ & $v_z^2$ & -1.0 \\
    Angular velocity $(x y)$ & $\|\boldsymbol{\omega}_{x y}\|_2^2$ & -0.5 \\
    Orientation & $\exp \left\{- \frac{\|\mathbf{g}_{x y}\|_2^2}{\sigma_g} \right\}$ & 1.0 \\
    Body height & $\exp \left\{- \frac{\left(h^{\text {target}}-h\right)^2}{\sigma_h} \right\}$ & 1.0 \\
    Action rate & $\|\mathbf{a}_t-\mathbf{a}_{t-1}\|_2^2$ & -2.0 \\
    Hip roll joint deviation & $\exp \left\{ - \frac{\theta_{\text{hip roll}}^{2}}{\sigma_{\text{roll}}} \right\}$ & 0.5 \\
    Hip yaw joint deviation & $\exp \left\{ - \frac{\theta_{\text{hip yaw}}^{2}}{\sigma_{\text{yaw}}} \right\}$ & 0.5 \\ 
    Ankle joint deviation & $\exp \left\{ - \frac{\theta_{\text{ankle pitch}}^{2}}{\sigma_{\text{pitch}}} \right\}$ & 0.5 \\ 
    Joint position limits & $\sum\limits_{\text{i}}\text{RELU}(\theta_{\text{i}} - \theta^{\max}_{\text{i}})$ & -5.0 \\
    Motor torque limits & $\sum\limits_{\text{i}}\text{RELU}(\tau_{\text{i}} - \tau^{\max}_{\text{i}})$ & -1e-3 \\
    Feet slip & $\sum\limits_{\text{f}}\left|\mathbf{v}_\text{f}\right| \cdot \mathds{1}_{\text {contact, f}}$ & -1.0 \\ 
    Feet air time & $\sum\limits_{\text{f}}\left|\mathbf{t}_{\text {airtime, f}} -0.5\right| \cdot \mathds{1}_{\text {touchdown, f}}$ & 0.5 \\
    \bottomrule
    \end{tabular}
    \label{tab:rewards}
\end{table}

The training process is illustrated in Fig. \ref{fig:train_scheme}. Tables \ref{tab:rewards}, \ref{tab:ppo}, and \ref{tab:domain_randomization} present the reward terms, algorithm parameters, and domain randomization settings, respectively.

\begin{table}[htp]
\centering
\caption{Overview of Domain Randomization.}
\label{tab:domain_randomization}
\begin{tabular}{llll}
\toprule
\textbf{Parameter} & \textbf{Unit} & \textbf{Range} & \textbf{Operator} \\
\midrule
Friction with ground & - & [0.75, 1.25] & scale \\
Link masses & - & [0.9, 1.1] & scale \\
Push velocity & m/s & [0.0, 2.5] & add \\
Linear velocity & m/s & [-0.1, 0.1] & add \\
Angular velocity & rad/s & [-0.2, 0.2] & add \\
Motor positions & rad & [-0.01, 0.01] & add \\
Motor velocities & rad/s & [-1.5, 1.5] & add \\
\bottomrule
\end{tabular}
\end{table}

\subsection{Robustness enhancement techniques}
While the core training methodology provides a foundation for learning locomotion policies, achieving real-world robustness requires additional techniques. To enhance the robustness and natural appearance of our locomotion control policy, we implemented three key techniques:

\textbf{Symmetry loss:} Following approaches in \cite{long2024learning}, \cite{symmetry}, and \cite{xie2025humanoid}, we incorporate a symmetry-inspired loss that promotes natural and symmetric gait patterns. This is implemented by adding the following term to the Actor-Critic loss: 
\begin{equation}
    \begin{aligned}
    \mathcal{L}_{\text{symmetry}} &= MSE(S_a(\pi(\mathbf{o}_t)), \pi(S_o(\mathbf{o}_t))\\
    \end{aligned}
    \label{eq:simmetry}
\end{equation}
where $S_a$ and $S_o$ map actions and observations to their symmetric equivalents along the $XZ$ plane. This approach not only increases sample efficiency but also leads to improved overall reward.

\textbf{Weight decay:} To prevent the policy from generating extreme motor position targets, we regularize the last layer of the actor network with the following loss term:
\begin{equation}
    \begin{aligned}
\mathcal{L}_{\text{decay}} = \lambda ||\theta_{-1, \text{actor}}||^2
    \end{aligned}
    \label{eq:decay}
\end{equation}

The complete loss function combines these components:
\begin{equation}
\mathcal{L} = \mathcal{L}_{\text{actor}} + \mathcal{L}_{\text{critic}} + \mathcal{L}_{\text{symmetry}} + \mathcal{L}_{\text{decay}}
\label{eq:loss_total}
\end{equation}

\textbf{Adversarial attacks:} To enhance robustness against observation uncertainties, external disturbances, and rapidly changing high-level commands, we implement adversarial training similar to \cite{adversarial}. The adversarial network generates perturbations to commanded velocities and the gravity vector, while also applying external forces to the robot's feet. Our training process follows three phases: 
\begin{itemize}
    \item \textbf{Phase 1}: Train only the locomotion policy for 1000 iterations
    \item \textbf{Phase 2}: Freeze the locomotion policy and train the adversarial network for 200 iterations
    \item \textbf{Phase 3}: Freeze the adversary network and train the locomotion policy for another 200 iterations
\end{itemize} 

\begin{figure*}[t]
   \centering
   \includegraphics[width=150mm]{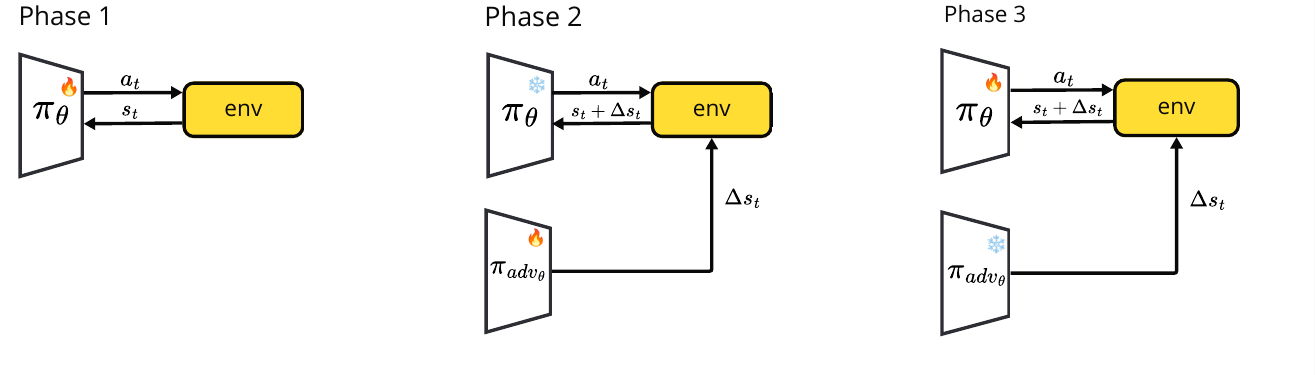}
   \caption{An overview of our cyclic adversarial training. \textbf{Phase 1}: training the base policy. \textbf{Phase 2}: training adversarial policy. \textbf{Phase 3}: training base policy under perturbations of adversarial policy.}
   \label{fig:method}
\end{figure*}

We repeat Phases 2 and 3 five times to progressively improve robustness. Fig. \ref{fig:method} provides an overview of this cyclic training process. This iterative approach allows the policy to adapt to increasingly challenging perturbations, resulting in more robust behavior in real-world scenarios.

\section{Experimental Validation}

\begin{figure}
    \centerline{\includegraphics[scale=0.18]{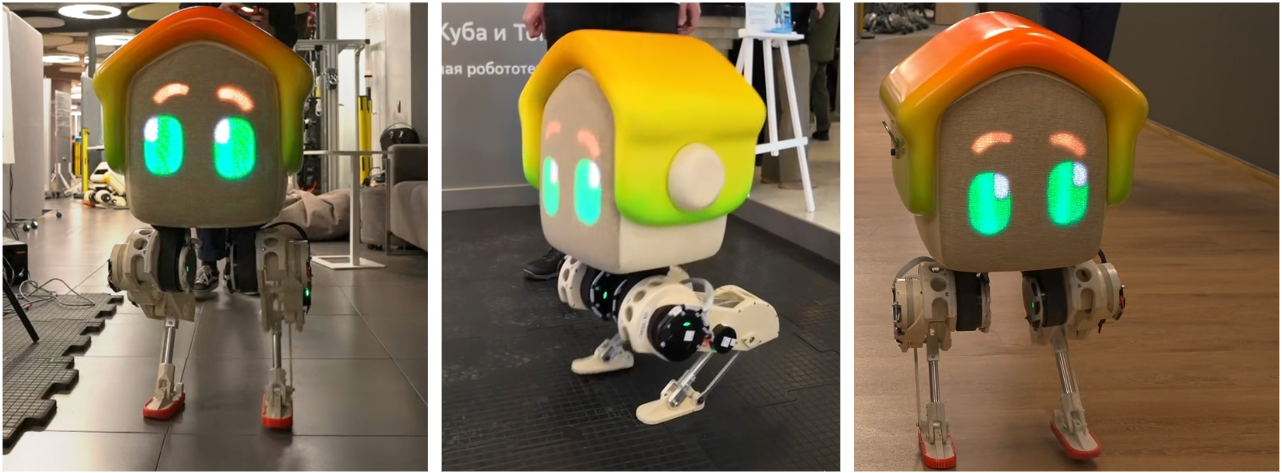}}
    \caption{The TopA walks in real world.}
    \label{fig:cheba_walking}
    \vspace*{-5mm} 
\end{figure}

With our learning methodology established, we now turn to experimental validation to assess its effectiveness. This section first introduces our custom bipedal platform, TopA, which features the closed kinematic chains that motivated our approach. We then present a series of experiments designed to systematically evaluate the performance of our method compared to conventional approaches, both in simulation and on the physical robot.

For our experimental platform, we utilize TopA (Fig. \ref{fig:cheba_depiction_live}), a bipedal robot with 5-DOF legs featuring coupled knee and ankle joints that form closed kinematic chains. Standing 70 cm tall and weighing 20 kg, TopA provides an ideal testbed for studying the challenges of parallel mechanisms in locomotion control. Its design exemplifies the practical constraints that often necessitate closed-loop linkages in bipedal robots.


\subsection{Hardware platform}
TopA is primarily constructed from 3D-printed plastic components, with aluminum rods providing structural reinforcement. The robot features 5-DOF legs, including hip roll, hip yaw, hip pitch, knee pitch, and ankle pitch joints. A key feature is the coupling between knee and ankle joints, which creates a motor-to-joint mapping:
\begin{equation}
\begin{bmatrix}
     q_{k} \\
     q_{a} \\
\end{bmatrix}_{j} = 
\begin{bmatrix}
     1 & 0 \\
     -1 & 1 \\
\end{bmatrix}
\begin{bmatrix}
     q_{k} \\
     q_{a} \\
\end{bmatrix}_m
\end{equation}

where $k$, $a$, $m$, and $j$ represent knee, ankle, motor, and joint, respectively. While this specific coupling can be represented linearly, our methodology works with virtually any holonomic constraint produced by closed-loop mechanisms. The robot's feet are covered with rubber soles to ensure good traction.

The robot's sensing and computational capabilities are designed to support our reinforcement learning approach. It is equipped with off-the-shelf T-MOTOR and ENCOS actuators and an Xsens IMU. To minimize the sim-to-real gap, we carefully identified the mechanical parameters of these actuators, including inertia, torque constant, and friction models (both viscous and dry). This identification was performed by applying frequency-rich signals to the motors and matching the MuJoCo models to the resulting data. We also incorporated these friction models into our IsaacLab learning pipeline.

For linear velocity estimation, we implemented a contact-aided Linear Kalman Filter (LKF) similar to that described in \cite{mit-cheetah}. This filter fuses IMU data with joint kinematics and runs automatically as part of the robot's software stack. We optimized the LKF parameters by comparing its estimates against ground truth data from an OptiTrack Motion Capture system.

The control policy $\pi$ operates at 50 Hz and provides position commands for all actuators of the robot: $\mathbf{a}_t \sim q_{\text{cmd}}$ as well as other robot  software runs on an onboard PC (NUC 12 Pro Kit with Intel Core i7-1270P) and maintains low-level communication at 400 Hz using ROS2 \cite{ros2}. These position commands are converted to torque commands with a PD controller operating at several kHz on the motor driver level. 

\subsection{Ablation study}
To rigorously evaluate the effectiveness of our approach, we conducted a comprehensive ablation study that isolates and quantifies the contribution of each component. Our analysis focuses on three distinct policy configurations:

\begin{itemize}
    \item \textbf{O:} Baseline policy trained with an open-loop kinematics model
    \item \textbf{C:} Enhanced policy utilizing our closed-loop kinematics model
    \item \textbf{C+A:} Complete policy combining closed-loop kinematics with adversarial training
\end{itemize}

To ensure a fair comparison, all policies were trained with identical reward structures (Tab. \ref{tab:rewards}), domain randomization parameters (Tab. \ref{tab:domain_randomization}), weight decay regularization, and symmetry considerations. We evaluated these policies in the MuJoCo simulator under conditions that closely mirror our IsaacSim training environment. Table \ref{tracking} presents our velocity tracking results across various locomotion tasks.

\begin{table}[h]
\centering
\caption{Commanded velocity tracking performance}
\begin{tabular}{ccccc}
    \toprule
     \multicolumn{5}{c}{Walk forward: $v_x=0.5$ m/s} \\
    \midrule
    \textbf{Policy} & \textbf{$v_{x,y},w_z$ MAE} & \textbf{$\bar{\tau}$ N/m} & \textbf{$\bar{\theta}_{roll}$} & \textbf{$\bar{\theta}_{pitch}$} \\
    \midrule
O & 0.58 & 35.90 & -0.01 & 0.22\\
C$^*$ & 0.76 & 23.18 & -0.01 & 0.81\\
C+A & 0.34 & 27.04 & 0.00 & 0.12\\
    \midrule
     \multicolumn{5}{c}{Walk backwards: $v_x=-0.35$ m/s} \\
    \midrule
     \textbf{Policy} & \textbf{$v_{x,y},w_z$ MAE} & \textbf{$\bar{\tau}$ N/m} & \textbf{$\bar{\theta}_{roll}$} & \textbf{$\bar{\theta}_{pitch}$} \\
     \midrule
O & 0.56 & 31.20 & -0.01 & -0.07\\
C & 0.47 & 28.46 & -0.01 & -0.25\\
C+A & 0.52 & 31.69 & -0.02 & -0.21\\
    \midrule
 \multicolumn{5}{c}{Walk right: $v_y=0.5$ m/s} \\
    \midrule
     \textbf{Policy} & \textbf{$v_{x,y},w_z$ MAE} & \textbf{$\bar{\tau}$ N/m} & \textbf{$\bar{\theta}_{roll}$} & \textbf{$\bar{\theta}_{pitch}$}\\
     \midrule
O & 0.43 & 33.31 & -0.02 & -0.02\\
C & 0.33 & 25.91 & -0.03 & -0.12\\
C+A & 0.23 & 28.49 & -0.02 & 0.01\\
     \midrule
  \multicolumn{5}{c}{Walk left: $v_y=-0.5$ m/s} \\
    \midrule
     \textbf{Policy} & \textbf{$v_{x,y},w_z$ MAE} & \textbf{$\bar{\tau}$ N/m} & \textbf{$\bar{\theta}_{roll}$} & \textbf{$\bar{\theta}_{pitch}$} \\
     \midrule
O & 0.43 & 33.94 & 0.02 & -0.03\\
C & 0.34 & 25.85 & 0.03 & -0.12\\
C+A & 0.28 & 28.98 & 0.02 & 0.04\\
     \midrule
  \multicolumn{5}{c}{Rotate counterclockwise: $w_z=1.0$ rad/s} \\
    \midrule
     \textbf{Policy} & \textbf{$v_{x,y},w_z$ MAE} & \textbf{$\bar{\tau}$ N/m} & \textbf{$\bar{\theta}_{roll}$} & \textbf{$\bar{\theta}_{pitch}$}\\
     \midrule
O$^*$ & 1.60 & 35.02 & 0.28 & -0.88\\
C & 0.35 & 23.38 & -0.01 & -0.08\\
C+A & 0.30 & 25.53 & -0.01 & 0.04\\
     \midrule
      \multicolumn{5}{c}{Rotate clockwise: $w_z=-1.0$ rad/s} \\
    \midrule
     \textbf{Policy} & \textbf{$v_{x,y},w_z$ MAE} & \textbf{$\bar{\tau}$ N/m} & \textbf{$\bar{\theta}_{roll}$} & \textbf{$\bar{\theta}_{pitch}$} \\
     \midrule
O$^*$ & 1.39 & 31.19 & 0.77 & -1.23\\
C & 0.35 & 23.38 & -0.02 & -0.09\\
C+A & 0.26 & 24.64 & 0.01 & -0.00\\
    \bottomrule 
    
     \multicolumn{5}{l}{$*$ Failure due to fall} \\
\end{tabular}
\label{tracking}
\end{table}

        It is clear that the tracking performance of the controller trained with the proposed methodology provides a substantial increase in velocity tracking performance across all axes compared to the policy trained using an open-loop kinematics model. These results highlight the importance of accurate modeling for effective policy learning.
        
        \subsection{Robustness evaluation}
        Beyond basic tracking performance, we conducted additional experiments to assess the robustness of our approach under various challenging conditions. These tests are particularly important for evaluating real-world applicability, where disturbances and unexpected inputs are common.

\begin{figure}
    \begin{subfigure}[b]{0.5\textwidth}
    \centering
    \scalebox{0.55}{\input{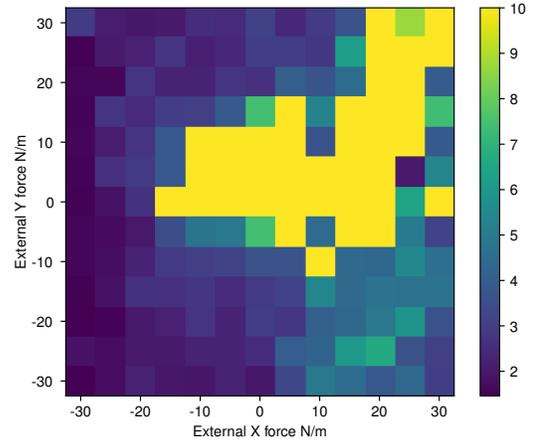}}
    \caption{Open: $\bar{t}_{fall}=4.77$}
    \label{fig:vanilla_noattack}
    \end{subfigure} 

    \begin{subfigure}[b]{0.5\textwidth}
    \centering
    \scalebox{0.55}{\input{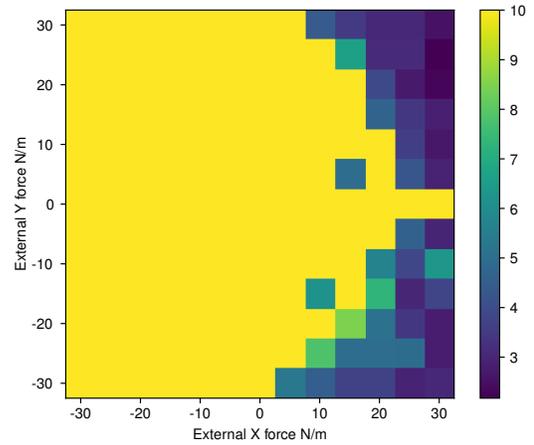}}
    \caption{Closed without adversary: $\bar{t}_{fall}=8.48$}
    \label{fig:noattack}
    \end{subfigure}
    
    \begin{subfigure}[b]{0.5\textwidth}
    \centering
    \scalebox{.55}{\input{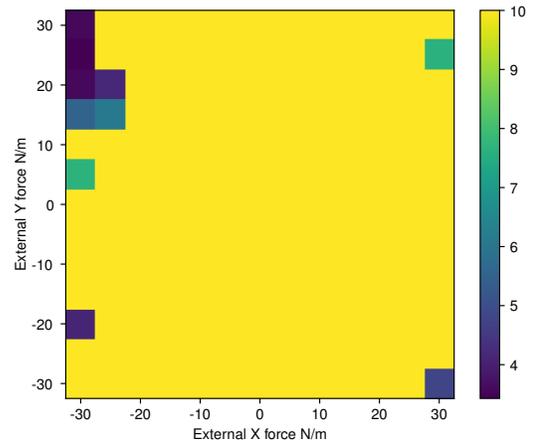}}
    \caption{With adversary: $\bar{t}_{fall}=9.71$}
    \label{fig:attack}
    \end{subfigure}
    \caption{Average time to fall under applied external force. All trained policies utilize weight decay regularization and symmetry loss.}
    \label{fig:heatmaps}
    \end{figure}

        \textbf{Stability under external perturbations:} To assess robustness against external forces, we applied constant disturbances to the torso of robot in its frame. We systematically varied both the direction and magnitude of these forces (from -30 N to 30 N in 5 N increments) and measured the time until failure over a 10-second window with results shown in Fig. \ref{fig:heatmaps}.
    
 The heatmaps in Fig. \ref{fig:heatmaps} reveal several critical insights about our approach. First, the vanilla policy with open-loop kinematics (Fig. \ref{fig:vanilla_noattack}) demonstrates poor disturbance rejection capabilities, with an average fall time of only 4.77 seconds. Its stability region is narrow, particularly vulnerable to lateral forces and backward pushes.

 The closed-loop kinematics model without adversarial training (Fig. \ref{fig:noattack}) shows a dramatic improvement, with the average time to failure increasing to 8.48 seconds—a 78\% improvement over the vanilla policy. This confirms our hypothesis that accurate modeling of closed kinematic chains is essential for robust control.

 Finally, the policy trained with both closed-loop kinematics and adversarial training (Fig. \ref{fig:attack}) achieves the best performance with an average fall time of 9.71 seconds, a 14\% improvement over the closed-loop model alone and a 104\% improvement over the vanilla policy. Most notably, this policy exhibits a more symmetric and expanded stability region, particularly in the forward direction, indicating enhanced ability to maintain balance under a wider range of disturbances.

 The adversarially trained policy also shows a more balanced response to forces from different directions, suggesting that it has learned more generalized strategies for maintaining stability rather than being optimized for specific disturbance patterns. This generalization capability is crucial for real-world deployment where the robot will encounter unpredictable external forces.

    \textbf{Robustness to command variability:} Building on our perturbation analysis, we also evaluated how well each policy handles unpredictable user commands by introducing random perturbations to the velocity commands while the robot attempted to walk forward at 0.5 m/s. We gradually increased the perturbation variance from 0.1 to 0.7 m/s for linear velocity and from 0.5 to 2.0 rad/s for angular velocity. Each configuration was tested over 100 runs, with results shown in Table \ref{user_input_robustness}.

    \begin{table}[]
        \centering
        \begin{tabular}{ccc}
        \toprule
        \textbf{$\sigma_v$ m/s} & \textbf{$\sigma_{w_z}$ rad/s} & \textbf{Success Rate (\%)} \\
        \midrule
        \multicolumn{3}{c}{Policy C} \\
        \midrule
        0.175 & 0.5 & 100 \\
        0.35 & 1.0 & 79 \\
        0.525 & 1.5 & 34\\
        0.7 & 2.0 & 12 \\
        \midrule
        \multicolumn{3}{c}{Policy C+A} \\
        \midrule
        0.175 & 0.5 & 98  \\
        0.35 & 1.0 & 99 \\
        0.525 & 1.5 & 85 \\
        0.7 & 2.0 & 73\\
        \bottomrule
        \end{tabular}
        \caption{Robustness evaluation under varying commanded velocity noise ranges}
        \label{user_input_robustness}
       \vspace{-4mm}
    \end{table}

    As can be seen from the Table \ref{user_input_robustness} the policy trained without adversarial perturbations (\textbf{C}) rapidly loses effectiveness as command noise increases, with success rates dropping to just 12\% at the highest noise levels. In contrast, our adversarially trained policy (\textbf{C+A}) maintains a 73\% success rate even under severe command perturbations, demonstrating substantially better resilience to unpredictable inputs. This significant improvement validates our hypothesis that adversarial training is essential for handling the variability encountered in real-world operation.

    \subsection{Real-world validation}
    The validation of our approach comes from its performance on the physical robot in real-world conditions (Fig.~\ref{fig:cheba_walking}). This phase represents the most stringent test of our methodology, where the full complexity of the sim-to-real gap becomes apparent.

    We conducted extensive testing of our trained policies on the physical TopA robot in an indoor facility. The test course spanned approximately 100 meters and incorporated various challenging elements: different surface materials (tiles and wood), glass doorframes, and narrow corridors requiring precise navigation.

    The results clearly demonstrated the superiority of our approach. The policy trained using the open-loop kinematics model (\textbf{O}) failed to complete the course, exhibiting poor balance control and frequent joint limit violations. This outcome underscores the critical importance of accurate simulation modeling. While both closed-loop policies successfully navigated the entire course, their performance differed significantly: the policy without adversarial training (\textbf{C}) required approximately 6 minutes to complete the course, whereas our complete approach (\textbf{C+A}) finished in just 3.1 minutes—demonstrating nearly twice the efficiency. Moreover, the \textbf{C+A} policy exhibited exceptional reliability during public demonstrations, maintaining continuous operation for 40 minutes without any interventions. These real-world results provide compelling evidence for the effectiveness of our integrated approach to bipedal locomotion control.

\section{Conclusion}

he experimental results presented in this paper demonstrate that accurate modeling of closed kinematic chains is essential for effective reinforcement learning in bipedal locomotion. By faithfully representing parallel mechanisms rather than defaulting to simplified serial chains, we achieved substantial performance improvements—including up to 81\% better tracking accuracy while requiring significantly lower torque inputs. Our comprehensive robustness framework, which integrates symmetry-aware learning, adversarial training, and policy regularization, proved particularly effective in handling unpredictable real-world conditions. Under severe command perturbations, our approach maintained a 73\% success rate, compared to just 12\% for the baseline implementation.

Real-world validation strongly supported these findings: our approach enabled the robot to navigate the test course in less than half the time required by the non-adversarial policy, while the simplified open-loop model proved incapable of completing the course. These results establish that proper modeling of closed kinematic chains is not merely a theoretical consideration but a practical necessity for bridging the sim-to-real gap. Looking ahead, we plan to explore domain adaptation techniques and linear velocity-free observation spaces to further reduce the system's dependence on precise system identification and state estimation. This work lays the foundation for more robust and adaptable bipedal locomotion control in real-world applications.

\bibliographystyle{IEEEtran}

\bibliography{ref}
\end{document}